\pdfoutput=1

\documentclass[11pt]{article}

\usepackage[final]{acl}

\usepackage{times}
\usepackage{latexsym}
\usepackage{booktabs}
\usepackage{url}

\usepackage[T1]{fontenc}

\usepackage[utf8]{inputenc}

\usepackage{microtype}

\usepackage{inconsolata}

\usepackage{graphicx}

\usepackage[utf8]{inputenc}
\usepackage{tcolorbox}

\usepackage{caption}
\usepackage{enumitem}
\usepackage{lipsum}
\usepackage{subcaption} 

\title{Using Language Models to Disambiguate Lexical Choices in Translation}

\author{
 \textbf{Josh Barua}\hspace{0.5cm}
 \textbf{Sanjay Subramanian}\hspace{0.5cm}
 \textbf{Kayo Yin}\hspace{0.5cm}
 \textbf{Alane Suhr}
\\
\\
University of California, Berkeley
\\
\texttt{\{joshbarua,sanjayss,kayoyin,suhr\}@berkeley.edu}
}

\begin{document}
\maketitle

\begin{abstract}

In translation, a concept represented by a single word in a source language can have multiple variations in a target language.
The task of lexical selection requires using context to identify which variation is most appropriate for a source text.
We work with native speakers of nine languages to create DTAiLS, a dataset of 1,377 sentence pairs that exhibit cross-lingual concept variation when translating from English.
We evaluate recent LLMs and neural machine translation systems on DTAiLS, with the best-performing model, GPT-4, achieving from 67 to 85\% accuracy across languages.
Finally, we use language models to generate English rules describing target-language concept variations.
Providing weaker models with high-quality lexical rules improves accuracy substantially, in some cases reaching or outperforming GPT-4. 

\end{abstract}

\section{Introduction}

Resolving ambiguity in translation is a fundamental challenge~\cite{weaver-1952-translation} that remains unsolved \cite{campolungo-etal-2022-dibimt}. 
This paper focuses on \textit{lexical selection}, a key aspect of translation that requires using context in a source sentence to determine the best translation for an ambiguous source word from several target-language options.
Figure~\ref{fig:rule_example} shows variations of the concept \textit{date} (fruit) in Farsi and an example of the lexical selection task.

\begin{figure}[ht]
\fbox{\parbox{0.95\columnwidth}{
    \scriptsize 
    \texttt{\noindent \textbf{\underline{Concept}:} \textit{date} (fruit) \\ \\
    \textbf{\underline{Variations and Generated Rules}} \\
    \textbf{\textit{Khorma}} refers to the fruit of the date palm when it is fully ripe and dried. It is commonly consumed as a sweet, chewy snack or used in various dishes, particularly desserts. \\\\
    \textbf{\textit{Rotab}} refers to fresh, soft dates that are partially ripe. These dates are moister and sweeter than fully ripe, dried dates (khorma). Rotab is often eaten as a fresh fruit or used in cooking where a softer, sweeter texture is desired. \\\\
    \textbf{\textit{Kharak}} refers to dates that are unripe and are in a semi-dried state. They are less sweet compared to rotab and khorma and are often used in cooking or further processed into other forms. \\ \\
    \textbf{\underline{Lexical Selection}} \\
    \textbf{Source Sentence:} she brings me dried \underline{dates} and tells me old stories \\
    \textbf{Correct Variation:} khorma
    }
}}
\caption{Generated rules for English \textit{date} with lexical variations \textit{khorma}, \textit{rotab}, and \textit{kharak} in Farsi.
}
\vspace{-1em}
\label{fig:rule_example}
\end{figure}

Our work has two main goals. First, we investigate the capabilities of language models in disambiguating lexical choices in translation by comparing instruction-tuned language models with high-performing neural machine translation systems. 
Second, we test whether language models can be used to extract useful natural language rules that accurately describe how to translate ambiguous words based on source-side context.

We work with native speakers to introduce the Dataset of Translations with Ambiguity in Lexical Selection (DTAiLS), a test set of 1,377 sentence pairs spanning nine languages where concept variation can be explained by context in the source sentence.
Evaluating five models on DTAiLS reveals that, without rules provided in-context, only the best-performing LLM, GPT-4, is competitive with NMT systems.

We also present a simple method for generating rules for lexical selection from LLMs, which native speakers verify are highly accurate. 
Figure~\ref{fig:rule_example} shows rules generated for three Farsi variations of the concept \textit{date}.
We observe improvements in performance across all LLMs when prompted with accurate self-generated rules.
In addition, while open-weight LLMs lag behind both NMT systems and GPT-4, providing rules from GPT-4 can help substantially to bridge the gap in performance.
This suggests that parametric knowledge of concept variation poses a greater challenge to models than the ability to apply such knowledge in-context.
Our work demonstrates that LMs can generate high-quality rules, and further leverage these rules to rival specialized NMT systems on lexical selection, but still fall short of native speakers.\footnote{Code and data are publicly released here: \url{https://github.com/Berkeley-NLP/Lex-Rules}}





\section{Task and Data}
To evaluate a model's ability to understand concept variations, we study lexical selection in translation.
For example, the noun \textit{date} has multiple lexical variations in Farsi, which distinguishes variations by fruit ripeness and dryness (Figure~\ref{fig:rule_example}).
We collect a dataset of translation pairs that require understanding and appropriately applying concept variation, where sufficient context is provided to distinguish between variations.

\paragraph{Lexical Selection}
In translation, lexical selection is the task of selecting the most appropriate lexeme in the target language that maps from a single lexeme in the source language, in the context of a source sentence~\cite{apidianaki-2009-data}.
Formally, let $(\bar{x}, \bar{y})$ be a sentence pair where $\bar{x} = \langle{x_1, \ldots, x_{|\bar{x}|}}\rangle$ is a sequence of words in the source language and $\bar{y} = \langle{y_1, \ldots, y_{|\bar{y}|}}\rangle$ is its translation in the target language. 
For a source word $x_i$, we define the set of possible translations of $x_i$ as $\bar{v} = \langle{v_1, \ldots, v_{|\bar{v}|}}\rangle$ where $\exists\; j$ such that $v_j \in \bar{y}$. 
The task of lexical selection is to identify the most appropriate translation $v_j$ conditioned on the source sentence $\bar{x}$.

\paragraph{Source Data}
Despite the existence of large-scale datasets for low-resource languages through bitext mining techniques \cite{schwenk-etal-2021-ccmatrix}, we focus on datasets curated by human translators to mitigate the potential for incorrect translations due to misalignment. 
We use OpenSubtitles \cite{lison-tiedemann-2016-opensubtitles2016, lison-etal-2018-opensubtitles2018}, TED2020 \cite{reimers-gurevych-2020-making}, PMIndia \cite{Haddow2020PMIndiaA}, and TEP \cite{10.1007/978-3-642-19437-5_6} to acquire parallel data for English paired with 7 low-resource and 2 high-resource languages (Japanese and Farsi).\footnote{Table~\ref{table:parallel_corpora} lists the data sources and number of sentence pairs available per language.}
All datasets are downloaded from the digital platform OPUS\footnote{\url{https://opus.nlpl.eu/}}~\cite{Tiedemann2009NewsFO}.

\paragraph{Expert Recruitment}
We work with bilingual speakers to ensure our methods and data faithfully represent the processes associated with translation under concept variation.
For each language, we recruit from Prolific\footnote{\url{https://www.prolific.com/}} three annotators who are fluent English speakers and native speakers of the target language.
All annotators are paid \$16 USD / hour.\footnote{More details on annotator recruitment and task design are available in Appendix~\ref{app:experts}.}

\begin{table}[h!]
  \centering\footnotesize
  \begin{tabular}{lccc}
    \toprule
    \textbf{Language} & \textbf{\# Concepts} & \textbf{Precision} & \textbf{Recall} \\
    & \textbf{Extracted} & \\
    \midrule
    Afrikaans & \phantom{0}17 & 99.2 & 82.4 \\
    Armenian & \phantom{0}18 & 85.4 & 77.8 \\
    Farsi & 100 & 96.2 & 72.3 \\
    Galician & \phantom{0}24 & 95.8 &  91.7 \\
    Hindi & \phantom{0}41 & 96.2 &  89.4 \\
    Japanese & 202 & 97.6 & 78.9 \\
    Latvian & \phantom{0}16 & 99.1 & 87.5 \\
    Tamil & \phantom{0}18 & 92.4 & 79.6 \\
    Telugu & \phantom{0}21 & 95.9 & 87.3 \\
    \bottomrule
  \end{tabular}
  \caption{Details for identifying concepts with variations, including the number of extracted concepts and the precision and recall of the extracted variations. On average, we identify 2.3 variations per concept.
  \vspace{-1em}
  }
  \label{table:parallel_sentences}
\end{table}

\subsection{Identifying Concepts with Variations}\label{sec:concepts}
We first identify concepts that are represented as a single word in our source language (English) but have several variations in a target language.
We build upon \citet{chaudhary-etal-2021-wall}'s approach to identify fine-grained lexical distinctions that arise due to concept variation.
Given a parallel corpus, we lemmatize all words using Stanza \cite{qi-etal-2020-stanza} and compute word alignments for each sentence pair with the AWESOME aligner \cite{dou-neubig-2021-word}.
Using these alignments, we create a one-to-many mapping from source-language lexemes to target-language lexemes.
Lastly, we remove source lexemes that do not map to at least two target lexemes, exhibit low entropy, or correspond to target lexemes that arise due to polysemy.\footnote{See Appendix~\ref{app:mining} for additional details on the pipeline.}
While this approach was originally designed and applied to Spanish and Greek parallel corpora, we apply it to nine additional languages.
Table~\ref{table:parallel_sentences} lists the total number of extracted concepts for each language.

\begin{table*}[h!]
  \footnotesize\centering
  \begin{minipage}[t]{0.7\textwidth}
    \centering
    \begin{tabular}{lcccc}
      \toprule
      \textbf{Language} & \textbf{Full Dataset} & \textbf{\% Sentences} & \textbf{Expert Dataset} \\
      & & \textbf{w/ Sufficient Context} \\
      \midrule
      Afrikaans & \phantom{0}4,123 & 78.9 & 180\\
      Armenian & \phantom{0}9,610 & 63.9 &176\\
      Farsi & 43,911 & 59.0 & 127\\
      Galician & 13,393 & 67.0 & 164\\
      Hindi & 17,417 & 61.0 & 145\\
      Japanese & 99,741 & 76.0 & 149\\
      Latvian & \phantom{0}6,944 & 81.3 & 184\\
      Tamil & \phantom{0}5,833 & 65.3 & 134\\
      Telugu & \phantom{0}9,167 & 63.3 & 118\\
      \bottomrule
    \end{tabular}
    \caption{DTAiLS dataset statistics. A sentence is determined to have sufficient context when the majority of annotators select its ground truth variation.}
    \label{table:lexical_selection_data}
  \end{minipage}%
  \hfill
  \begin{minipage}[t]{0.25\textwidth}
    \centering
    \begin{tabular}{lc}
      \toprule
      \textbf{Language} & \textbf{\% Correct} \\
    & \textbf{Rules} \\
      \midrule
      Afrikaans & \phantom{0}99.2 \\
      Armenian & \phantom{0}86.2 \\
      Farsi & \phantom{0}95.1 \\
      Galician & \phantom{0}99.4 \\
      Hindi & \phantom{0}98.5 \\
      Japanese & \phantom{0}99.1 \\
      Latvian & 100.0 \\
      Tamil & \phantom{0}99.2 \\
      Telugu & \phantom{0}98.0 \\
      \bottomrule
    \end{tabular}
    \caption{Mean accuracy of GPT-4 generated rules.}
    \vspace{-1em}
    \label{table:correct_rules}
  \end{minipage}
\end{table*}

We also perform comprehensive analysis of this approach's precision and recall in identifying target-language variations of concepts.
All three expert annotators for each language provide feedback on the extracted variations, including whether each variation matches the meaning of the English lexeme (for computing precision) and whether any key variations are missing from the set (for computing recall).
Precision is measured as the proportion of accurate variations; recall is measured as the proportion of concepts with all key variations recovered.\footnote{Appendix~\ref{app:annotation} includes more details on this analysis, including inter-annotator agreement.} 
In general, the precision of identified variations is very high, even for low-resource languages, but our approach is less consistent in identifying all possible variations of a source concept, which could be due to the limited size of datasets used or the use of domain-specific data.

\subsection{Dataset Construction}
Our goal is to collect a dataset of sentence pairs that require understanding target-language concept variation for accurate translation.
Expert annotators help us curate this dataset by performing the lexical selection task, provided only source-language sentences and target lexemes. 
All annotators for a given language are presented with the same set of concepts and source-language sentences.
We shuffle the order in which concepts, sentences, and target lexemes are shown to each annotator.
Our resulting dataset, DTAiLS, includes sentences for which the majority of annotators selected the variation used in the original sentence pair, which indicates that there is sufficient context for consistent lexical selection.
Although there could be cases of context-dependent translation where there isn't a single optimal lexical variation, for our dataset we rely on majority agreement to select examples that are clearly evaluable.
Table~\ref{table:lexical_selection_data} includes dataset statistics.\footnote{ Appendix~\ref{app:exampleselection} contains more dataset construction details.}



\section{Rules for Lexical Selection}\label{sec:lexicalrules}
We experiment with generating human-readable English rules that describe all extracted concepts and their variations, and analyze how these rules influence model performance on lexical selection when provided as input.\footnote{
We refer to these natural language descriptions as ``rules'' to be consistent with prior work studying lexical selection \cite{chaudhary-etal-2021-wall}.}

For each target-language variation, we find the 50 longest source-language sentences with less than 50 tokens in length where the variation appears in the target translation.
Motivated by work showing LLMs are useful for describing differences between sets of text~\cite{pmlr-v162-zhong22a}, we construct one prompt per concept including all target-language variations and their respective lists of source-language sentences, and prompt each model to provide a natural language description of each target-language variation.\footnote{
The Appendix (Figure~\ref{box3}) shows the prompt template.}
We generate rules from three instruction-tuned models: Gemma-1.1~\cite{gemmateam2024gemma}, LLaMA-3~\cite{llama3modelcard}, and GPT-4~\cite{openai2024gpt4};
Figure~\ref{fig:rule_example} provides an example rule set from GPT-4.
For each language, we ask all three native speakers to label every GPT-4-generated rule for correctness.
Table~\ref{table:correct_rules} shows these rules are overwhelmingly correct according to native speakers. 

\begin{figure*}[h!]
    \centering
    \includegraphics[width=0.9\textwidth]{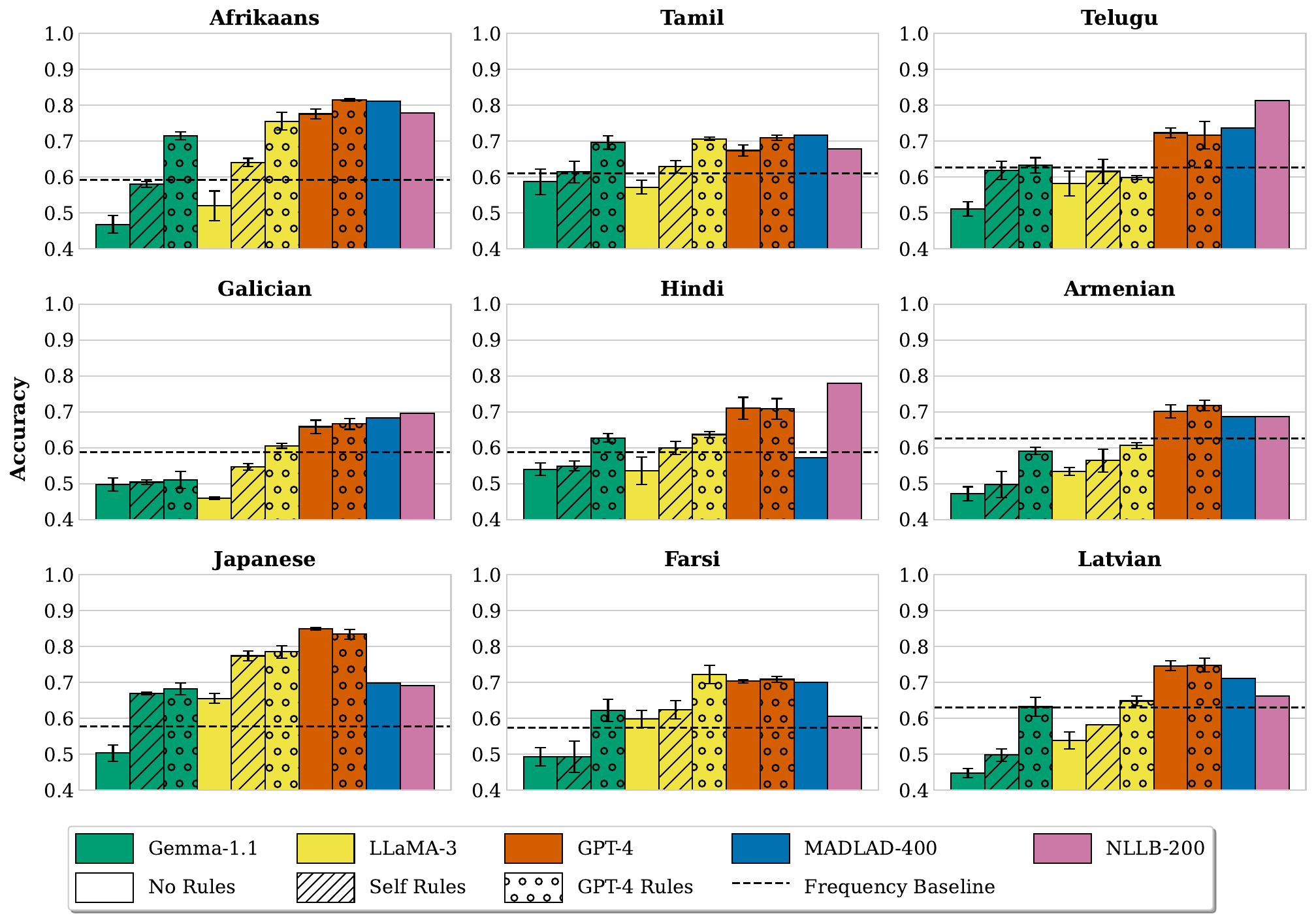}
    \caption{Comparisons between LMs with and without rules to NMT systems on lexical selection. We report $\mu_{\pm \sigma}$ across 3 runs for LM experiments.
    }
    \label{fig:results}
\end{figure*}

\section{Experiments}




We evaluate three instruction-tuned models, GPT-4, LLaMA-3-8B-Instruct, and Gemma-1.1-7B-IT; and two high-quality NMT models MADLAD-400-10B-MT~\cite{kudugunta2023madlad400} and NLLB-200-3.3B~\cite{team2022NoLL}, sampling with temperature of 0.

We hypothesize that models performing lexical selection will benefit from access to rules describing concept variations (Section~\ref{sec:lexicalrules}).
Thus, we evaluate instruction-tuned LLMs in 3 settings: (1) no rules provided, (2) with self-generated rules, and (3) with rules generated by GPT-4. 
For each setting, we instruct the model to explain its reasoning prior to selecting a lexical choice~\cite{Wei2022ChainOT}.\footnote{Appendix~\ref{app:experiments} includes prompts used for lexical selection and a description of lexical selection with NMT systems.}




Accuracy is measured as the proportion of sentences for which the model selects the correct lexical variation. 
We include a simple frequency baseline that always predicts the most common target-language variation in our dataset for each concept during lexical selection.
Figure~\ref{fig:results} shows results for each model. 

First, we find that SOTA NMT systems trained on our low-resource language pairs outperform all LLMs on Telugu and Hindi, while being comparable on Afrikaans, Galician, Tamil, and Farsi. 
For the remaining three languages, the best-performing LLM, achieves a 4-15\% absolute improvement in performance over the NMT systems. 
We find a large gap in performance between open- and closed-weight LLMs, with the frequency baseline outperforming open-weight LLMs without rules in seven languages.
LLaMA-3-8B-Instruct outperforms Gemma-1.1-7B-IT for all nine languages.

Self-generated lexical rules improve model performance on nearly all languages, with improvements being even more significant for the weaker open-weight models.
This suggests that models can better reason over lexical choices if their parametric knowledge of concept variation is explicitly included as context in the prompt.
While lexical selection with GPT-4 is significantly more expensive than with open-weight models, rule generation has a one-time cost. 
When providing the open-weight models with rules acquired from the strongest model (GPT-4), we see total improvements up to 23\% in accuracy, with these models performing close to or even exceeding GPT-4 on several languages.
However, even when these high-quality (Table~\ref{table:correct_rules}) rules are provided, there is still a significant gap to human performance.
We hypothesize that while the generated rules are accurate, they fail to enumerate all possible contextual factors that could influence lexical choice in all translation settings. 



\section{Related Work}


The most closely related work is by \citet{chaudhary-etal-2021-wall}, who study lexical selection for English to Spanish and Greek.
They present a pipeline for collecting lexical selection data, train models to perform lexical selection, use a linear SVM model to extract features as interpretable rules, and evaluate the efficacy of these rules in a second-language acquisition setting. 
In contrast, we use modern LMs, generate natural language rules, and evaluate on several low-resource languages.
We also curate a test set for lexical selection validated by native speakers of our target languages.

Lexical selection is closely related to the problem of ambiguity in machine translation, where context is essential for disambiguating various possible translations. 
\citet{fernandes23acl} investigate such ambiguity that arises from discourse and grammar, while \citet{campolungo-etal-2022-dibimt} explore ambiguity due to polysemy. 
\citet{iyer-etal-2023-towards} evaluate LLMs on  translations under polysemy,  demonstrating that in-context learning and fine-tuning on ambiguous datasets improves translation. 
We study the potential for LLMs to resolve ambiguity arising from target-language concept variation, focusing on low-resource languages.

Prior work has shown improvements in LLM translation quality by incorporating ground truth dictionary entries into prompts \cite{ghazvininejad2023dictionary}.
We further demonstrate that models can accurately describe concept variations in low-resource languages using only parametric knowledge and example usages from source-language sentences.
Our experiments follow a line of work showing that modern LLMs exhibit non-English language capabilities, though these LLMs are often trained primarily on English data~\cite{robinson-etal-2023-chatgpt,asai2023buffet}.



\section{Conclusion}
We introduce DTAiLS, a dataset of 1,377 sentence pairs with 9 language pairs that exhibit ambiguity in translation due to concept variation. 
Using this dataset, we evaluate 3 popular instruction-tuned LLMs and 2 high-performing NMT systems on the task of lexical selection.
Out of nine languages tested, the strongest LLM outperforms the NMT systems on three languages, has comparable performance on four languages, and fall short of these systems on two languages.
No model is able to disambiguate the full set of sentences that native speakers can.

We also present a simple approach to extract high-quality rules from language models, demonstrating improvements on lexical selection when LMs are given access to rules.
We find that providing weaker open-weight models with rules from a stronger LLM can effectively bridge the gap to or even outperform the stronger model for several languages.
Future research could investigate additional applications of lexical rules in NMT and assess how these human-readable rules can assist L2 learners in vocabulary acquisition.



\section*{Limitations}
Because our focus is on low-resource languages, the parallel corpora we use are small; thus, we are only able to extract roughly 20 concepts for six out of nine languages.
Further, due to the time and effort required to collect human judgements on lexical selection,\footnote{Annotating the full Japanese dataset would require roughly 5 thousand total hours of work.} our test sets curated by experts are just 120-180 examples per language and 1,377 examples overall.
Developing automated methods for example selection is an interesting direction for future work that will enable larger-scale evaluation. 
We also note that the recall of the pipeline for identifying concepts with variations might be inaccurate due to the challenges annotators face brainstorming all possible variations in the semantic space.
Lastly, due to a lack of available models for WSD, dependency parsing, and POS tagging for low-resource languages, we are only able to evaluate on language pairs where English is the source language.
In theory, the methods we present can work for any arbitrary language pair. 

\section*{Acknowledgments}
The authors are grateful to the anonymous reviewers for helpful feedback.
This work was partially supported by AI2 Young Investigator Grant and a Gemma Academic Program Award.
KY is supported by the Future of Life PhD Fellowship.
During this project, JB was supported by the Berkeley SURF program.
We would also like to thank Amartya Bose, Niloofar Mireshghallah, Prithviraj Ammanabrolu, Yanai Elazar, Mihran Miroyan, PV Subramanian, and Elzette van Rensburg for helpful discussions on generated lexical rules and dataset construction.
Finally, we are deeply thankful to all of the crowdworkers who participated in our study, without whom this work would not have been possible.

\bibliography{acl_latex}


\appendix

\clearpage

\section{Additional Annotation Details}\label{app:annotation}
\subsection{Expert Annotation}\label{app:experts}
\begin{figure*}[t]
    \centering
    \includegraphics[width=0.95\textwidth]{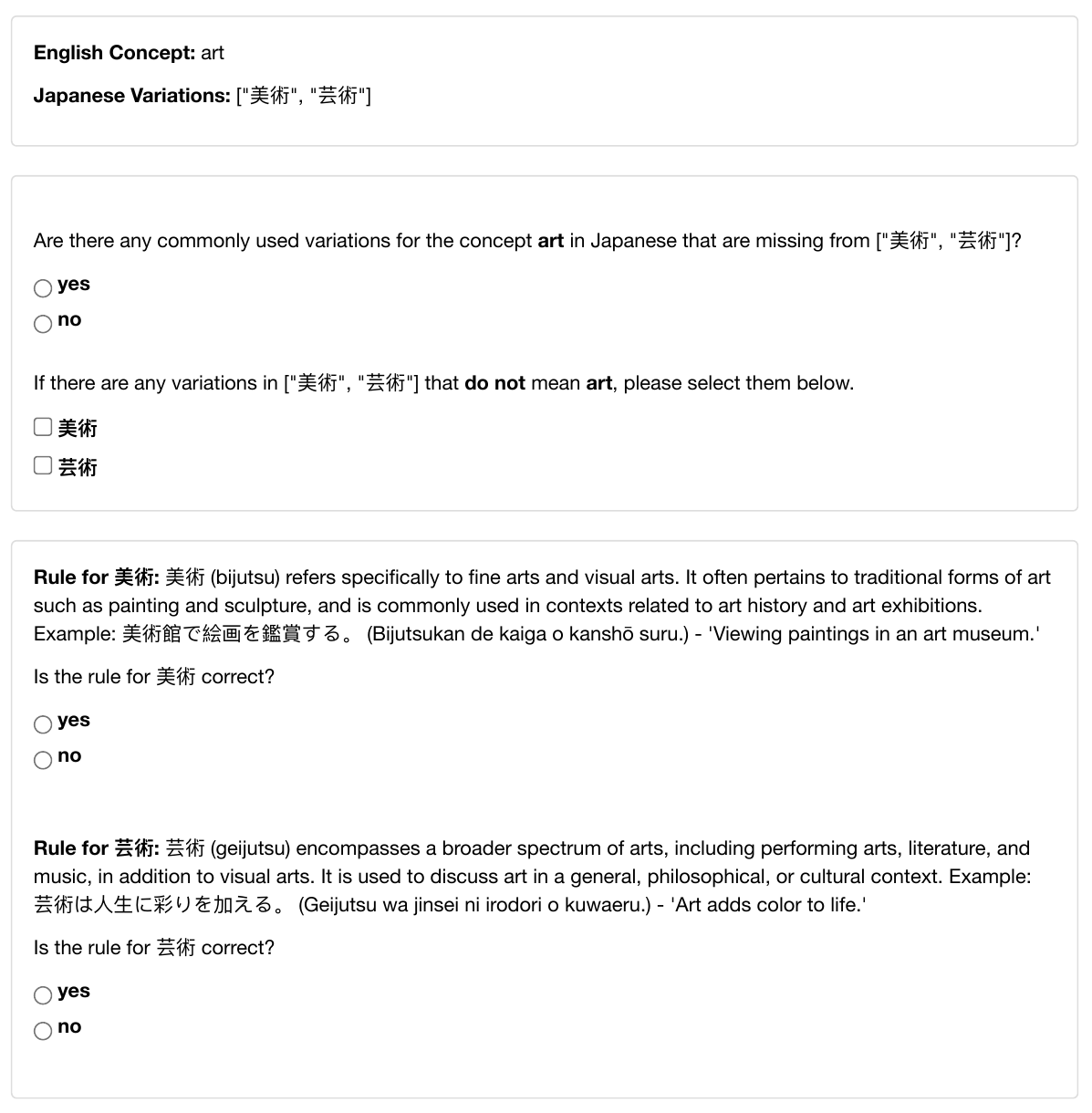}
    \caption{Interface for annotating rules and extracted concepts and variations.}
    \label{figure:rules_interface}
\end{figure*}

\begin{figure*}[t]
    \centering
    \includegraphics[width=0.95\textwidth]{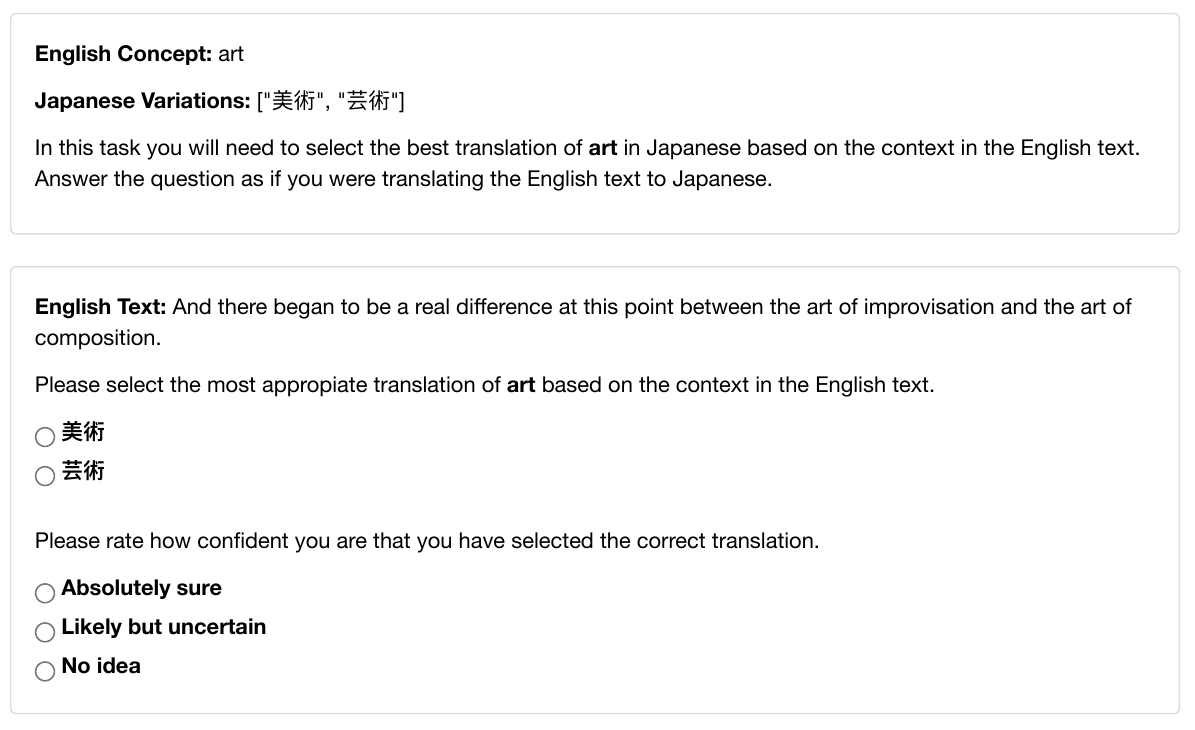}
    \caption{Interface for lexical selection task.}
    \label{figure:lexical_selection_interface}
\end{figure*}

The filters applied for participants before joining include (1) highest level of degree earned as technical/community college or above and (2) fluency in English and native proficiency in one of the nine languages we study.
First, a pilot study was conducted to vet annotators for fluency and comprehension of the task.
We then published two studies to the group of annotators who qualified from the pilot study.
The first study required annotation of the rules generated by GPT-4 and feedback on concepts and lexical variations extracted by the pipeline (Section~\ref{sec:concepts}) for computing precision and recall.
The interface for the first study can be found in Figure~\ref{figure:rules_interface}.
The second study required annotators to complete the lexical selection task.
The interface for the second study can be found in Figure~\ref{figure:lexical_selection_interface}.
We ensure that the same annotator doesn't participate in both studies.
We remove all personal identifying information from all data collected. Prior to taking part in any study, annotators were informed of the purpose of the study and how their data would be used.

\subsection{Example Selection}\label{app:exampleselection}

Collecting human judgments of lexical selection for all parallel sentences is infeasible; for example, fully annotating Japanese would require labeling 99,741 examples (Table~\ref{table:lexical_selection_data}) and roughly 5 thousand total hours of work.
Due to limited resources, we sample up to 20 concepts for each target language and gather 10 sentence pairs per concept, ensuring that each variation is represented at least once.
This results in a test set of up to 200 sentences per language.
For languages with more than 20 extracted concepts, we first filter for concepts that have a roughly uniform distribution over variations. 
Specifically, for each concept we compute the relative frequency of each lexical variation. 
Concepts are discarded if any individual variation deviates by more than 20\% from a uniform distribution.
After filtering, we uniformly sample 20 concepts to be included in the lexical selection task.

\subsection{Inter-Annotator Agreement}
\label{sec:appendix}

In this section we report statistics on inter-annotator agreement for all studies conducted with native speakers.
We include 3 metrics: \textit{total agreement} is the proportion of questions for which all 3 annotators were in agreement, \textit{Fleiss' kappa} \cite{fleiss1971measuring} is a popular metric for measuring reliability of agreement between more than 2 annotators, and \textit{prevalence-adjusted bias-adjusted kappa} (PABAK) \cite{BYRT1993423} is a modified kappa that addresses problems that arise from an imbalanced distribution of data over classes.
We include Fleiss' kappa for completeness, but note that it is often misleading due to the very high prevalence of class 1 (correct).
For example, in the first row of Table~\ref{table:IAA_generated_rules} we find that despite annotators being in total agreement for over 97\% of questions, the Fleiss' kappa measure suggests poor agreement.
This is a common paradox for measures of inter-annotator agreement that are ``chance-corrected.''
Table~\ref{table:IAA_generated_rules} presents inter-annotator agreement statistics for evaluation of generated rules, while Tables~\ref{table:IAA_precision} and~\ref{table:IAA_recall} display the same for precision and recall of the concept variation extraction pipeline.

\begin{table}
  \centering\footnotesize
  \begin{tabular}{llll}
    \hline
    \textbf{Language} & \textbf{Total Agreement} & \textbf{Fleiss' $\kappa$} & \textbf{PABAK} \\
    \hline
    Afrikaans & 0.975 & -0.008 & 0.950 \\
    Armenian & 0.707 & 0.181 & 0.415 \\
    Farsi & 0.857 & -0.021 & 0.713 \\
    Galician & 0.982 & -0.006 & 0.964 \\
    Hindi & 0.954 & -0.016 & 0.908 \\
    Japanese & 0.973 & -0.009 & 0.945 \\
    Latvian & 1.000 & NaN & 1.000 \\
    Tamil & 0.977 & -0.008 & 0.955 \\
    Telugu & 0.959 & 0.319 & 0.918 \\
    \hline
  \end{tabular}
  \caption{Inter-annotator agreement statistics for evaluation of generated rules.}
  \label{table:IAA_generated_rules}
\end{table}

\begin{table}
  \centering\footnotesize
  \begin{tabular}{llll}
    \hline
    \textbf{Language} & \textbf{Total Agreement} & \textbf{Fleiss' $\kappa$} & \textbf{PABAK} \\
    \hline
    Afrikaans & 0.975 & -0.008 & 0.950 \\
    Armenian & 0.634 & 0.024 & 0.268 \\
    Farsi & 0.917 & 0.241 & 0.835 \\
    Galician & 0.909 & 0.254 & 0.818 \\
    Hindi & 0.931 & 0.376 & 0.862 \\
    Japanese & 0.951 & 0.307 & 0.903 \\
    Latvian & 0.974 & -0.009 & 0.947 \\
    Tamil & 0.818 & 0.134 & 0.636 \\
    Telugu & 0.959 & 0.652 & 0.918 \\
    \hline
  \end{tabular}
  \caption{Inter-annotator agreement statistics for precision of concept variation pipeline.}
  \label{table:IAA_precision}
\end{table}

\begin{table}
  \centering\footnotesize
  \begin{tabular}{llll}
    \hline
    \textbf{Language} & \textbf{Total Agreement} & \textbf{Fleiss' $\kappa$} & \textbf{PABAK} \\
    \hline
    Afrikaans & 0.588 & 0.056 & 0.176 \\
    Armenian & 0.444 & -0.071 & -0.111 \\
    Farsi & 0.360 & -0.066 & -0.280 \\
    Galician & 0.792 & 0.091 & 0.583 \\
    Hindi & 0.683 & -0.118 & 0.366 \\
    Japanese & 0.510 & 0.019 & 0.020 \\
    Latvian & 0.625 & -0.143 & 0.250 \\
    Tamil & 0.611 & 0.201 & 0.222 \\
    Telugu & 0.667 & -0.002 & 0.333 \\
    \hline
  \end{tabular}
  \caption{Inter-annotator agreement statistics for recall of concept variation pipeline.}
  \label{table:IAA_recall}
\end{table}


\begin{table*}[h]
  \centering\footnotesize
  \begin{tabular}{llcccc}
    \toprule
    \textbf{Language} & \textbf{Data Source} & \textbf{\# Parallel Sentences} \\
    \midrule
    Afrikaans & OpenSubtitles & \phantom{0}44,703 \\
    Armenian & TED2020 & \phantom{0}37,122 \\
    Farsi & TED2020, TEP & 916,975 \\
    Galician & TED2020 & \phantom{0}34,385 \\
    Hindi & OpenSubtitles, TED2020 & 140,649 \\
    Japanese & TED2020 & 366,661 \\
    Latvian & TED2020 & \phantom{0}55,488 \\
    Tamil & OpenSubtitles, TED2020 & \phantom{0}43,741 \\
    Telugu & OpenSubtitles, PMIndia, TED2020 & \phantom{0}72,860 \\
    \bottomrule
  \end{tabular}
  \caption{Details for parallel corpora of all nine language included in our study, including data sources and the number of parallel sentences.
  }
  \label{table:parallel_corpora}
\end{table*}

\section{Additional Experimental Details}
\label{app:experiments}

\subsection{Pipeline for Identifying Concepts with Variations}\label{app:mining}
In this section, we formally describe the pipeline for identifying concepts with variations, which we adopt from \citet{chaudhary-etal-2021-wall}. Let $\mathcal{D} = \{(\bar{x}_1, \bar{y}_1), \ldots, (\bar{x}_{|\mathcal{D}|}, \bar{y}_{|\mathcal{D}|})\}$ be a parallel corpus where $(\bar{x}_i, \bar{y}_i)$ is a source- and target-language sentence pair.
For each sentence pair, we compute word alignments and lemmatize all words in $\bar{x}_i$ and $\bar{y}_i$ using the AWESOME aligner and Stanza respectively.
Furthermore, for source sentences only, we perform automatic part-of-speech (POS) tagging and dependency parsing using Stanza and word-sense disambiguation (WSD) using EWISER \cite{bevilacqua-navigli-2020-breaking}.
Source words are characterized by tuples of their lemmatized form and POS tag $\langle{l_x, t_x}\rangle$ to avoid conflating different meanings across POS tags.
First, we enumerate all word alignments across the corpus and create a one-to-many mapping from each source word to the lexical variations it is aligned with.
Second, we remove all source words that do not map to at least two lexical variations at least 50 times.
We require 50 occurrences for each variation to prevent incorrect translations being extracted due to noisy alignments.
Next, we describe a process for filtering out source words based on entropy.
For a given source word tuple $\langle{l_x, t_x}\rangle$ with lexical variations $\bar{v} = \langle{v_1, \ldots, v_{|\bar{v}|}}\rangle$, let $n_i$ be the number of occurrences of variations $v_i$.
We compute the conditional probability of each variation $v_i$ as $$p(v_i \mid l_x, t_x) = \frac{n_i}{\sum_{j=1}^{|\bar{v}|}n_j}$$ and the entropy of a source word tuple as $$H(l_x, t_x) = \sum_{j=1}^{|\bar{v}|}-p(v_j \mid l_x, t_x)log_{e}(p(v_j \mid l_x, t_x))$$
We remove all source word tuples with an entropy below 0.69.
Lastly, we remove lexical variations that arise due to polysemy in the source word.
In particular, since source words are only characterized by their lemmatized form and POS, it is possible that we extract variations that correspond to different senses of $l_x$.
For example, the source word tuple $<plane, noun>$ is translated in Spanish as \textit{avión} when referring to an aircraft and \textit{plano} when referring to a geometric plane.
When each variation is mapped to a source word, we store the word sense of $l_x$ as it is used in the source sentence.
This allows us to compute the most frequent word sense during post-processing and remove variations belonging to other word senses.

\subsection{Lexical Selection with NMT Systems}\label{app:nmt}

To perform lexical selection with NMT systems, we pass the source sentence as input and parse the translated text for the predicted lexical variation.
First, we check for an exact substring match in the translated text with all lexical variations. If an exact match is found, we take that to be the predicted variation.
If not, we tokenize the translated text with Stanza and computing the Levenshtein ratio \cite{Levenshtein1965BinaryCC} between every word and every lexical variation.
We identify the variation with the highest ratio to any word in the translated text for fuzzy matching.
If this ratio exceeds 0.7, we select it as the predicted variation; otherwise, we conclude that no variation is found and the prediction is labeled as incorrect.

\subsection{Prompts}

Since the Gemma family of models do not take a system prompt as input, we prepend the system prompt to the user prompt with the role \textit{user} for all experiments involving Gemma-1.1.
Figures~\ref{box1}, \ref{box2}, and \ref{box3} show the prompts that we use to evaluate LMs on lexical selection and generate rules.
For lexical selection with LLMs, we apply the same fuzzy matching scheme as Section~\ref{app:nmt} to match the generated answer to a target-language variation.
Qualitatively, the instruction-following capabilities of GPT-4 were greater than that of Llama-3 and Gemma-1.1.
If any LM failed to generate an answer according to the provided template, we append ``Please enclose your selected translation from {\color{blue}<Translations>} with 3 back ticks.'' to the prompt and resample once.
If the LM fails to follow the template a second time, the prediction is labeled as incorrect.

\tcbset{userstyle/.style={colback=gray!5!white, colframe=black!75!black, fonttitle=\bfseries, arc=2mm, boxrule=0.5mm, width=\columnwidth}}


\subsection{Position Bias in Lexical Selection}
\begin{figure}[t]
    \centering
    \begin{minipage}{0.48\textwidth}
        \centering
        \includegraphics[width=\linewidth]{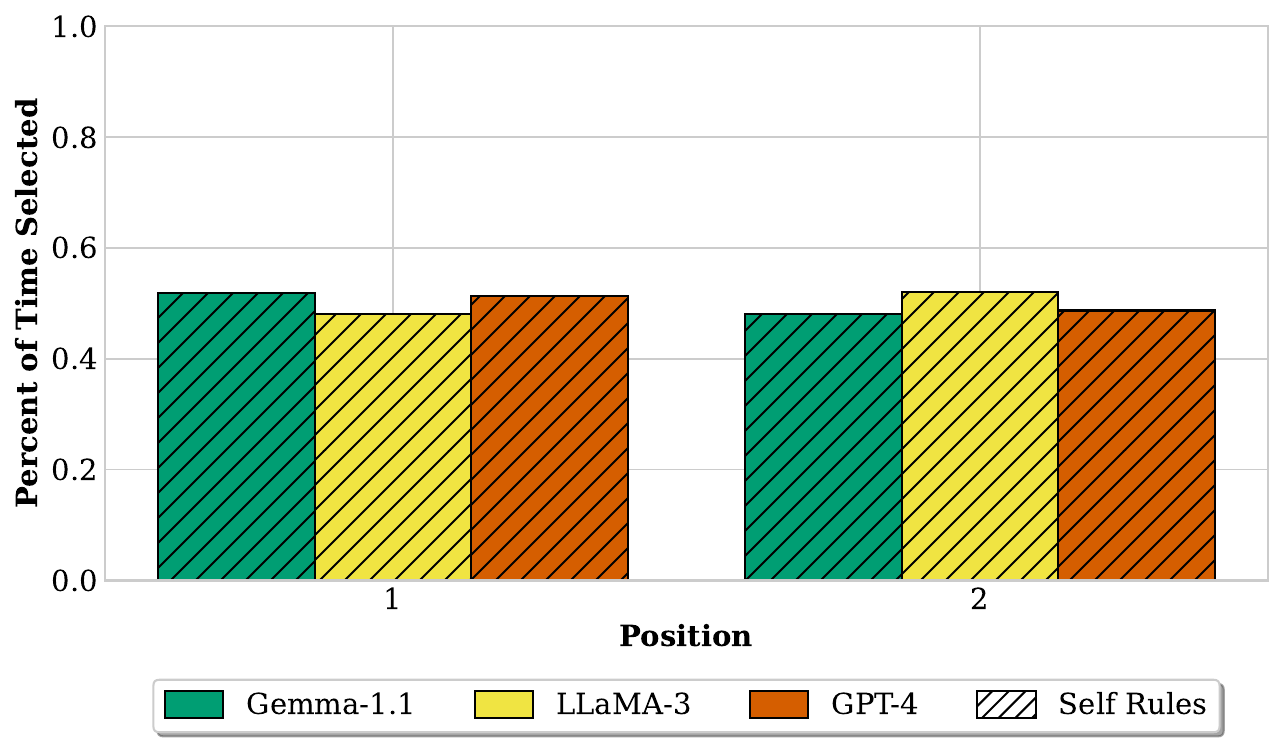}
        \caption{Percent of time each model selects an answer at each position when there are 2 lexical variations.}
        \label{figure:two_variations_bias}
    \end{minipage}\hfill 
    \vspace{0.05\textwidth}
    \begin{minipage}{0.48\textwidth}
        \centering
        \includegraphics[width=\linewidth]{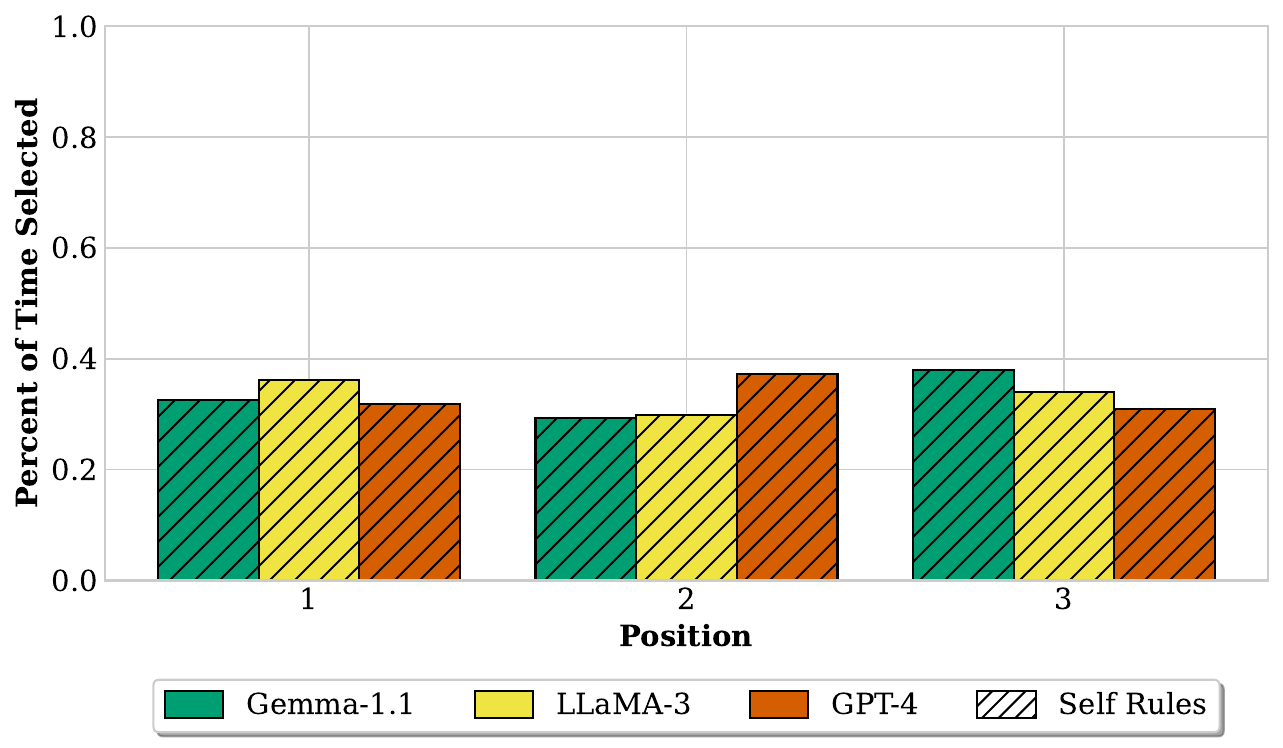}
        \caption{Percent of time each model selects an answer at each position when there are 3 lexical variations.}
        \label{figure:three_variations_bias}
    \end{minipage}
\end{figure}

We acknowledge that with our prompt, the lexical selection task is similar to a multiple choice question (MCQ) setup.
While humans tend to be order-invariant when answering MCQs, several prior works have examined position bias in LLMs when solving MCQs \cite{DBLP:conf/iclr/RobinsonW23, DBLP:journals/corr/abs-2309-03882}.
To ensure our evaluation is not affected by position bias we take three steps.
First, we shuffle the order of translations in the prompt for every example during lexical selection to reduce bias.
Second, we report how often the LMs select an answer choice at each position across the Afrikaans, Latvian, and Japanese subsets of DTAiLS.
Figures~\ref{figure:two_variations_bias} and ~\ref{figure:three_variations_bias} show a roughly uniform distribution over selected positions for concepts that have 2 and 3 lexical variations, respectively.
Lastly, we plot the mean and standard deviation of each LM experiment across 3 runs in Figure~\ref{fig:results}.
We find that the test accuracy is consistent for all models despite each run being initialized with a unique seed for shuffling the order of translations.
Based on these findings, we conclude that LMs are approximately order-invariant when doing lexical selection with our prompting setup.

\subsection{OpenAI Model Used}

In our call to the OpenAI API, we use the model name \textit{GPT-4-Turbo} which at the time of writing is a pointer to \textit{gpt-4-turbo-2024-04-09}.

\subsection{Computational Requirements}

All experiments in this paper that do not involve models from OpenAI require approximately 50 GPU hours on an NVIDIA RTX A6000 GPU.

\section{Software and Licenses}

The TED2020 dataset uses the CC-BY-NC-4.0 License. 
All models are utilized from Hugging Face; LLaMA-3-8B-Instruct uses the Llama3 License, Gemma-1.1-7B-IT uses the Gemma License, MADLAD-400-10B-MT uses the Apache License 2.0, and NLLB-200-3.3B uses the CC-BY-NC-4.0 License.
Our use of datasets and models is consistent with their intended use.

\begin{tcolorbox}[userstyle]
----------------- System Prompt ---------------- \\
------------------- User Prompt ------------------\\
Please select the best translation of "\textit{\color{blue}<Concept>}" in "\textit{\color{blue}<Source Text>}" from the following list: \textit{\color{blue}<Translations>}.
Carefully explain your reasoning first and then enclose your final answer like this {\textasciigrave\textasciigrave\textasciigrave}answer{\textasciigrave\textasciigrave\textasciigrave}.
\end{tcolorbox}
\captionof{figure}{Full prompt for the lexical selection task.}\label{box1}
\vspace{2mm}

\begin{tcolorbox}[userstyle]
----------------- System Prompt ---------------- \\
Here are rules for how to translate "\textit{\color{blue}<Concept>}" in \textit{\color{blue}<Target Language>}:\textit{\color{blue}<Rules>} \\
------------------- User Prompt ------------------\\
Based on the provided rules, please select the best translation of "\textit{\color{blue}<Concept>}" in "\textit{\color{blue}<Source Text>}" from the following list: \textit{\color{blue}<Translations>}.
Carefully explain your reasoning first and then enclose your final answer like this {\textasciigrave\textasciigrave\textasciigrave}answer{\textasciigrave\textasciigrave\textasciigrave}.
\end{tcolorbox}
\captionof{figure}{Full prompt for the lexical selection task with self-generated rules.}\label{box2}
\vspace{2mm}

\begin{tcolorbox}[userstyle]
----------------- System Prompt ---------------- \\
Please only return a json with the following keys \textit{\color{blue}<Translations>} and no other text.
For each key the value should be a string in English explaining how the meaning and usage of that \textit{\color{blue}<Target Language>} word is different from the others.
The string should also include a brief example in \textit{\color{blue}<Target Language>} of the word being used with an English translation.
Please include the transliteration from \textit{\color{blue}<Target Language>} to Latin characters if necessary. \\
------------------- User Prompt ------------------\\
When translating the concept "\textit{\color{blue}<Concept>}" from English to \textit{\color{blue}<Target Language>}, what is the difference in meaning between \textit{\color{blue}<Translations>} and in which contexts should they be used?
Here are sentences where each word is used in-context to help you: \textit{\color{blue}<Sentences>}
\end{tcolorbox}

\captionof{figure}{Full prompt for generating rules from LMs.}\label{box3}

\end{document}